\pdfoutput=1

\documentclass[11pt]{article}

\usepackage{acl}

\usepackage{times}
\usepackage{latexsym}

\usepackage[T1]{fontenc}

\usepackage[utf8]{inputenc}

\usepackage{microtype}
\usepackage{listings}
\usepackage{graphicx}
\usepackage{multirow}
\usepackage{tabularx}
 \usepackage{booktabs}

\title{Exploring Effectiveness of GPT-3 in Grammatical Error Correction: \\A Study on Performance and Controllability in Prompt-Based Methods}

\author{Mengsay Loem, Masahiro Kaneko, Sho Takase, and Naoaki Okazaki \\
         Tokyo Institute of Technology \\ 
         \texttt{\{mengsay.loem, masahiro.kaneko\}[at]nlp.c.titech.ac.jp} \\\texttt{sho.takase[at]linecorp.com}, \texttt{okazaki[at]c.titech.ac.jp}
}

\begin{document}
\maketitle
\begin{abstract}
Large-scale pre-trained language models such as GPT-3 have shown remarkable performance across various natural language processing tasks. 
However, applying prompt-based methods with GPT-3 for Grammatical Error Correction (GEC) tasks and their controllability remains underexplored. 
Controllability in GEC is crucial for real-world applications, particularly in educational settings, where the ability to tailor feedback according to learner levels and specific error types can significantly enhance the learning process.
This paper investigates the performance and controllability of prompt-based methods with GPT-3 for GEC tasks using zero-shot and few-shot setting. 
We explore the impact of task instructions and examples on GPT-3's output, focusing on controlling aspects such as minimal edits, fluency edits, and learner levels. 
Our findings demonstrate that GPT-3 could effectively perform GEC tasks, outperforming existing supervised and unsupervised approaches. 
We also showed that GPT-3 could achieve controllability when appropriate task instructions and examples are given.
\end{abstract}

\section{Introduction}
\label{sec:intro}

Grammatical Error Correction (GEC) is an essential application of Natural Language Processing (NLP) in educational settings, as it significantly enhances learners' language skills and writing performance~\cite{kaneko-etal-2022-interpretability}. 
In real-world applications, controlling specific GEC settings, such as minimal and fluency edits and learner level-based corrections, is crucial to address diverse learning needs and scenarios~\cite{napoles-etal-2017-jfleg,bryant-etal-2019-bea,flachs-etal-2020-grammatical}. 
Although recent GEC approaches based on supervised learning have achieved remarkable progress, they heavily rely on large training datasets comprising both genuine and pseudo data~\cite{xie-etal-2018-noising,ge-etal-2018-fluency,zhao-etal-2019-improving,lichtarge-etal-2019-corpora,xu-etal-2019-erroneous,choe-etal-2019-neural,qiu-etal-2019-improving,grundkiewicz-etal-2019-neural,kiyono-etal-2019-empirical,grundkiewicz-junczys-dowmunt-2019-minimally,wang-zheng-2020-improving,zhou-etal-2020-improving-grammatical,wan-etal-2020-improving,koyama-etal-2021-comparison}. 
Collecting such data for each specific setting is challenging and time-consuming, which limits the scalability of these methods in various learning situations.

Prompt-based methods utilize large-scale pre-trained language models (PLMs), such as GPT-3, and have demonstrated promising results in numerous NLP downstream tasks. 
These tasks include natural language inference, question answering, and summarization~\cite{brown-gpt3-2020, Radford2019LanguageMA}.
Given the demand for control in GEC tasks across various settings, prompt-based methods are appealing because they deliver exceptional performance without needing extensive labeled data.
Despite the success of prompt-based methods in multiple NLP tasks, their application to GEC remains under-explored.
Although \newcite{coyne2023analysis} and \newcite{fang2023chatgpt} have recently assessed prompt-based methods on select GEC benchmarks, a comprehensive analysis has yet to be conducted.
This study aims to bridge this gap by concentrating on in-depth analyses of prompt-based methods and their controllability, aspects that have not been thoroughly investigated in previous research.

Our research seeks to address the following questions: 
1) To what extent can PLMs using prompt-based methods solve GEC tasks? and 
2) Is it possible to control GEC settings with prompts written in natural language using prompt-based methods?

In this work, we demonstrate that prompt-based methods with GPT-3~\cite{brown-gpt3-2020} achieve outstanding performance in GEC tasks (Section \ref{sec:general-performance}). In addition, the approach provides better control over the GEC process using task instructions and examples (Section \ref{sec:controllability}). 
We conduct analyses to examine the impact of different types of task instructions on GPT-3's performance in both zero-shot and few-shot setting, which emphasizing the importance of appropriate task instructions for GEC tasks (Section \ref{exp-instruction-selection}). 
Additionally, we investigate the effect of varying the number of examples in few-shot setting, and reveal that performance improves as the number of examples increases, albeit not strictly linearly (Section \ref{exp-num-examples}).

Furthermore, we explore the model's controllability in various GEC scenarios, more specifically, its ability to concentrate on either minimal or fluency aspects (Section \ref{exp-control-fluency}) and edits based on learner levels (Section \ref{exp-learner-level}).
Experimental results indicate that task instructions alone may be sufficient to control editing without examples. 
However, we found that combining task instructions with examples resulted in more effective controlling performance. 
This indicates the importance of both task instruction and examples for better control of GEC settings using prompt-based methods, although the example set tends to have more importance.

\section{Overall Experimental Settings}
\label{exp-overall-settings}

In this study, we designed a series of experiments using the prompt-based method with GPT-3 to evaluate the performance in GEC tasks. 
We utilized the GPT-3 model (\texttt{text-davinci-003}) through the API provided by OpenAI\footnote{https://openai.com/blog/openai-api}. 
Our experiments were conducted in two settings: zero-shot and few-shot.

\paragraph{Zero-shot}
In the zero-shot setting, we assessed GPT-3's ability to perform GEC tasks without any prior examples. 
We employed the following template for prompts in the zero-shot setting:
\begin{verbatim}
{task instruction}: {input text};
output:___
\end{verbatim}

\paragraph{Few-shot}
For the few-shot setting, we implemented in-context learning as described by \newcite{brown-gpt3-2020}. 
We provided the model with a few examples to guide its understanding of the GEC task. 
We randomly sampled pairs of examples from the training (or validation) sets of each experimental setting to serve as examples for the model. 
Details on the number and source of examples used in each experiment are described in the corresponding sections below.
The template for prompts in the few-shot setting is as follows:
\begin{verbatim}
{task instruction}
{example 1}
...
{example N}
{input text}; output:___
\end{verbatim}

\paragraph{Prompt}
We used natural language text prompts for all our experiments.
The task instruction within the prompt serves as a directive that informs the model about the desired outcome of each task.
We varied the task instructions in both zero-shot and few-shot setting to examine the model's adaptability to different phrasings (refer to Section \ref{exp-instruction-selection}).
The instruction candidates employed in our prompt analyses are listed in Appendix~\ref{sec:appendix-instruction}.
Examples of task instructions include: 
\texttt{Correct the grammatical errors in the following sentence}, 
\texttt{Revise mistakes in this text}, and 
\texttt{Rewrite the following text with proper grammar}.

\begin{table*}[t]
\centering
\begin{tabular}{lccc}
\hline
\textbf{Method} & \textbf{JFLEG} & \textbf{CoNLL2014} & \textbf{W{\&}I+LOCNESS} \\ \hline
Transformer (big) & 53.22 & 51.11 & 51.36 \\
\hline
\newcite{grundkiewicz-junczys-dowmunt-2019-minimally} & 56.18 & 44.23 & 47.89 \\
\newcite{grundkiewicz-etal-2019-neural} & -- & 26.76 & -- \\
\hline
ChatGPT zero-shot with CoT~\cite{fang2023chatgpt} & 61.40 & 51.70 & 36.10 \\
GPT-3 zero-shot & 64.51 & 56.05 & 53.07 \\
GPT-3 16-shot & \textbf{67.02} & \textbf{57.06} & \textbf{57.41} \\
\hline
\end{tabular}
\caption{Comparison of GPT-3's performance using both supervised and unsupervised approaches on the JFLEG, CoNLL2014, and W{\&}I+LOCNESS test sets in zero-shot and few-shot settings, with 16 examples. The upper block of the table shows the results for the supervised approach, while the middle block shows the results for the unsupervised approaches. The scores are GLEU scores for JFLEG, ${\rm F}_{0.5}$ scores for CoNLL2014, and W{\&}I+LOCNESS.}
\label{table:results-overall}
\end{table*}

\section{General Performance}
\label{sec:general-performance}

To address research question 1) mentioned in Section~\ref{sec:intro}, we investigated the overall performance of the prompt-based method with GPT-3 in GEC tasks. 
This investigation is particularly relevant given the increasing prevalence of GPT-3 in various NLP applications and the need to assess its potential capabilities for GEC tasks specifically.

\subsection{Settings}

We evaluated the performance of GPT-3 on three GEC test sets: JFLEG~\cite{napoles-etal-2017-jfleg}, CoNLL2014~\cite{ng-etal-2014-conll}, and W{\&}I+LOCNESS~\cite{bryant-etal-2019-bea,Granger1998TheCL} using both zero-shot and few-shot settings with 16 examples. 
We used examples from the training set of JFLEG, NUCLE~\cite{dahlmeier-etal-2013-building}, and W{\&}I+LOCNESS as examples in the few-shot setting when evaluating with JFLEG, CoNLL2014, and W{\&}I+LOCNESS test sets, respectively.

We compared our prompt-based methods to baselines, including supervised and unsupervised approaches. 
For the supervised approach, we trained a Transformer (big) using the settings described in \newcite{vaswani-transformer-2017} and employed annotated data from multiple training sets. 
These sets included W{\&}I+LOCNESS, FCE corpus~\cite{yannakoudakis-etal-2011-new}, Lang-8 Corpus of Learner English~\cite{mizumoto-etal-2012-effect}, and NUCLE.
After removing uncorrected sentence pairs, the training data used to train the Transformer model was approximately 600K pairs. 
For unsupervised approach, we compared our methods to previous work in the literature including \newcite{grundkiewicz-junczys-dowmunt-2019-minimally} and \newcite{grundkiewicz-etal-2019-neural} where models were pre-trained with synthetic data.
We also compared with the result of ChatGPT performance in zero-shot with chain-of-thought (CoT) reported in \newcite{fang2023chatgpt}.

\subsection{Results}

Table~\ref{table:results-overall} shows the GLEU scores for JFLEG, ${\rm F}_{0.5}$ scores for CoNLL2014, and W{\&}I+LOCNESS. 
From the table, GPT-3 performed competitively in the GEC tasks in both zero-shot and few-shot settings, outperforming the Transformer model in all test sets. 
In the zero-shot setting, GPT-3 surpassed the Transformer, with gains of about 11, 5, and 2 percentage points on JFLEG, CoNLL2014, and W{\&}I+LOCNESS, respectively. 
The few-shot setting with 16 examples further improved GPT-3's performance, indicating the model's capability to adapt to the task with minimal examples quickly.

When comparing GPT-3 to unsupervised methods, we observe that GPT-3 outperforms other approaches in all test sets consistently. 
This comparison demonstrates the advantage of GPT-3 over existing unsupervised methods, even in the zero-shot setting.
When comparing the performance of ChatGPT in the zero-shot setting with CoT, GPT-3 outperforms ChatGPT CoT in all three test sets.
These results indicate GPT-3 is a more effective model for GEC tasks, especially in unsupervised settings.

\section{Investigation on Prompt}
\label{sec:investigate-prompt}
 
In this section, we analyze the impact of different factors in prompt on the performance of GPT-3 in GEC tasks. 
We focus on two factors: (1) the type of task instructions used and (2) the number of examples used in the few-shot settings.
Our primary objective is to comprehend the influence of various factors in prompts to the models' output, which will enable us to optimize GPT-3 more effectively for GEC tasks.
 
\subsection{Effect of Task Instruction}
\label{exp-instruction-selection}

In this section, we examine the effect of various types of task instructions on GPT-3's performance in GEC tasks.
We conduct evaluations using different task instructions in both zero-shot and few-shot settings.

\subsubsection{Settings}

We created three types of task instructions, with ten candidates per type, following related work on natural language inference task~\cite{webson-pavlick-2022-prompt}. 
The types of task instructions are as follows (See Appendix~\ref{sec:appendix-instruction} for details).
We used the JFLEG validation set in this experiment. 

\paragraph{Instructive} instructions explicitly request the model to correct grammatical errors in the given text, such as \texttt{Correct grammatical errors in this sentence} and \texttt{Revise grammatical mistakes in the following text}.

\paragraph{Misleading} instructions do not directly ask for grammar correction but instead require paraphrasing or rewriting, such as \texttt{Paraphrase the following sentence} and \texttt{Rewrite the following text to make it clearer}.

\paragraph{Irrelevant} instructions are unrelated to grammar correction, such as \texttt{Translate the following sentence} and \texttt{Write a news headline about this sentence}.

\subsubsection{Result}

\begin{figure}[t]
    \centering
    \includegraphics[width=\linewidth]{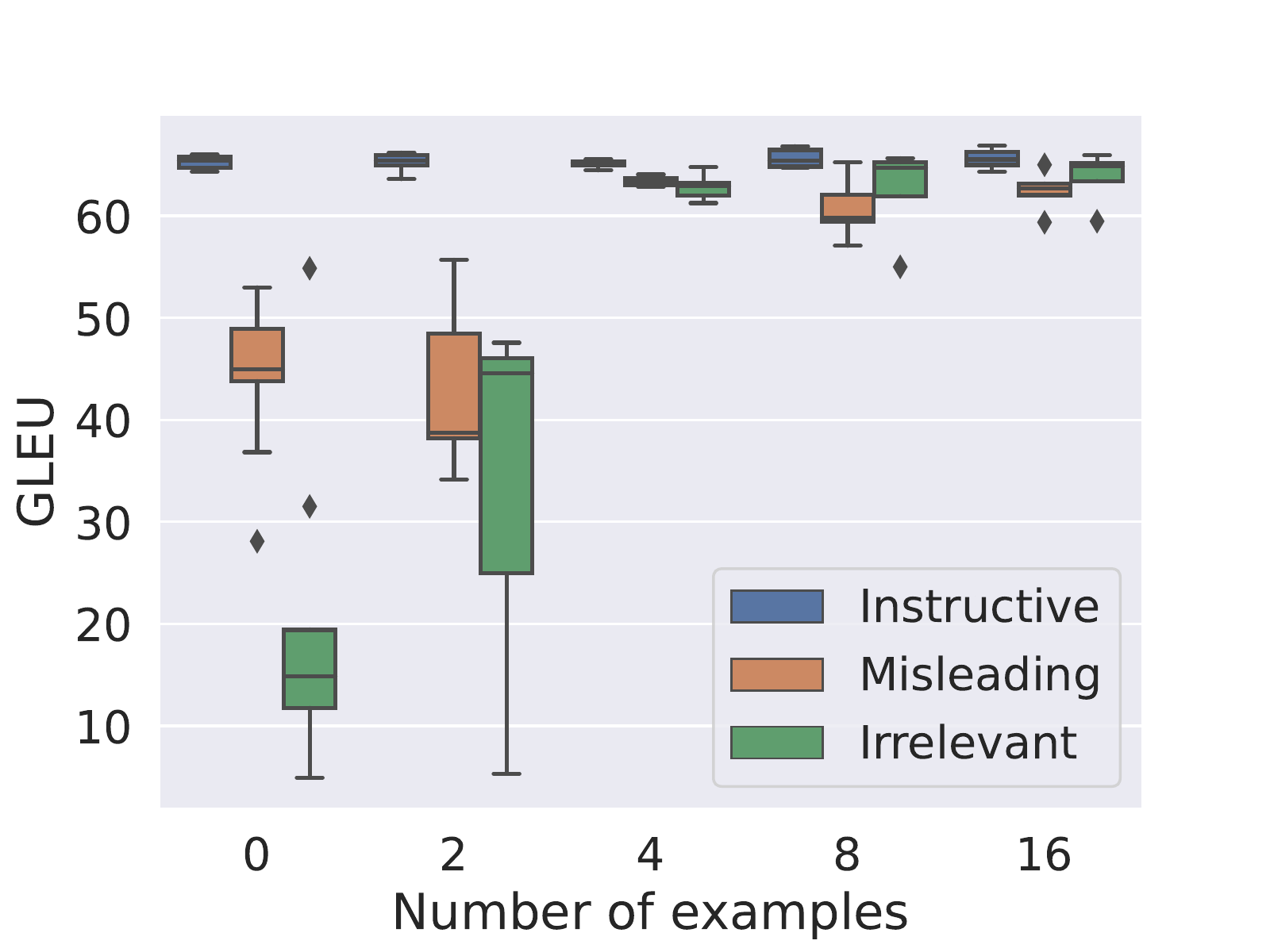}
    \caption{Comparison of GPT-3's performance using different types of task instructions (Instructive, Misleading, and Irrelevant) in zero-shot and few-shot settings on GEC tasks.}
    \label{fig:task-instruction}
\end{figure}

Figure~\ref{fig:task-instruction} shows the summary of the results when using different types of instructions in both zero-shot and few-shot settings. 
The findings reveal that task instructions significantly affect the performance of GPT-3 in GEC tasks. 

In the zero-shot setting, instructive instructions produced the highest average score (65.54), while irrelevant instructions resulted in the lowest average score (17.05), clearly demonstrating that the type of task instruction impacts the model's performance. 
Misleading instructions fell in the middle, with an average score of 43.45.

In few-shot settings, instructive instructions still outperformed the other two types, but the performance gap between instructive and misleading instructions decreased as the number of examples increased. 
The variance of the scores decreased with an increasing number of examples, suggesting that the model's performance becomes more consistent as it receives more examples. 

When comparing the different few-shot settings, we observed a clear trend of increasing performance as the number of examples increased. 
The standard deviation also decreased as the number of examples increased, indicating that the model's performance became more consistent with more examples.

\begin{figure}[t]
    \centering
    \includegraphics[width=\linewidth]{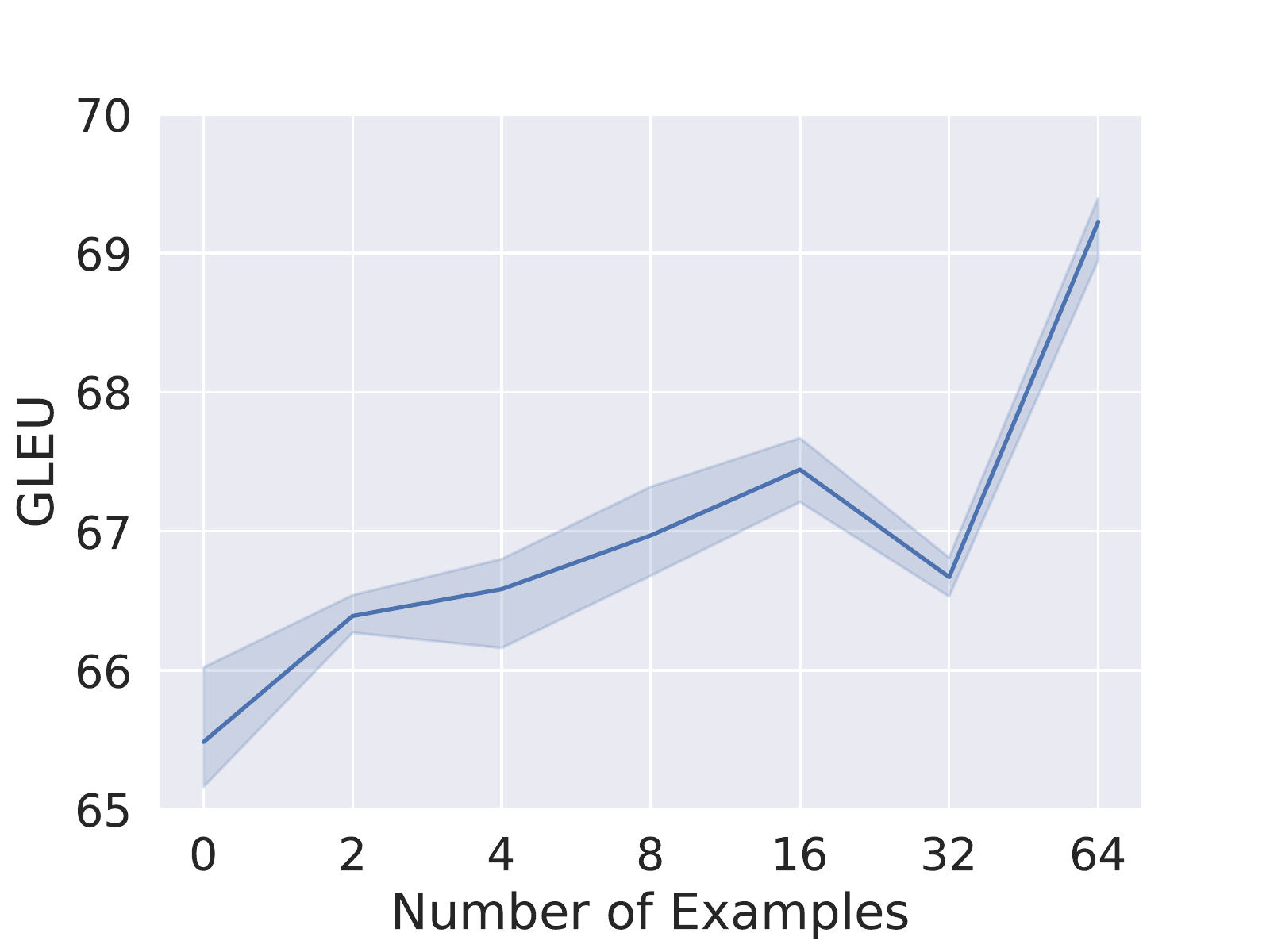}
    \caption{Effect of the number of examples on GPT-3's performance in few-shot settings, evaluated on the JFLEG test set with a fixed task instruction.}
    \label{fig:num-example}
\end{figure}

\begin{figure*}[t]
    \centering
    \includegraphics[width=\textwidth]{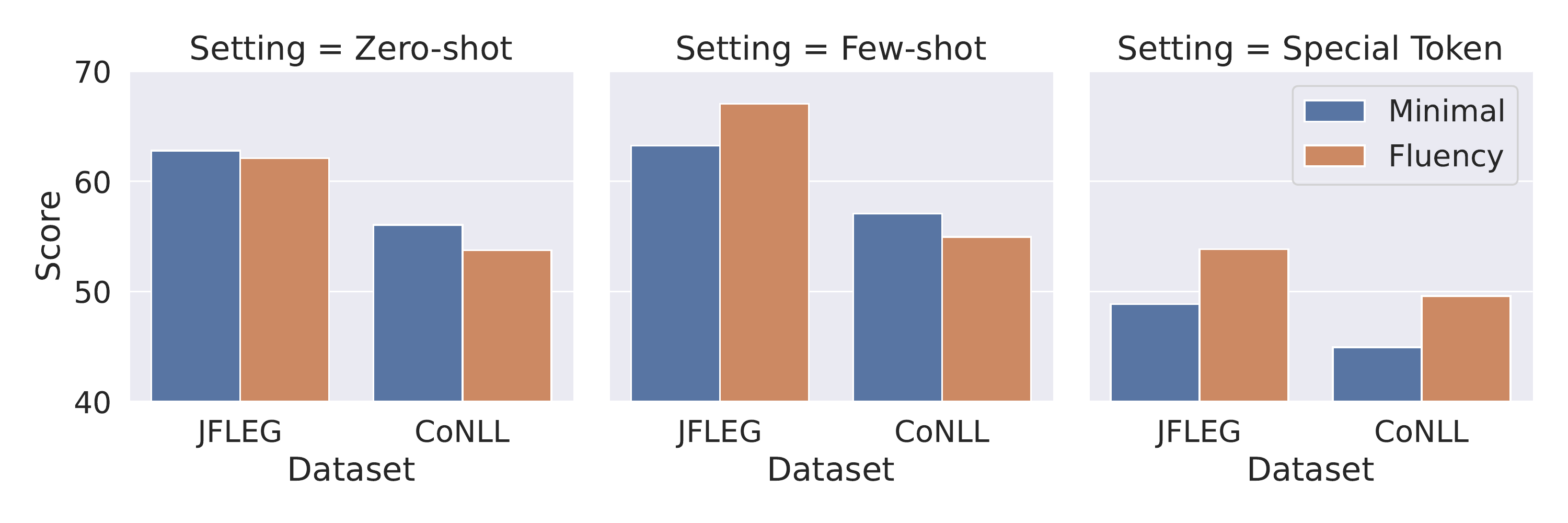}
    \caption{Comparison of GPT-3's controllability for minimal and fluency edits using CoNLL2014 and JFLEG test sets, respectively, measured in GLEU scores.}
    \label{fig:control-fluency-performance-all}
\end{figure*}

\subsection{Effect of Number of Examples}
\label{exp-num-examples}

In this section, we examine the impact of the number of examples used in few-shot settings on GPT-3's performance. 
Our objective is to understand how providing varying numbers of examples to the model influences its performance. 
By maintaining a fixed instruction and focusing solely on varying the number of examples, we aim to better comprehend their effect on the model's performance.

\subsubsection{Settings}

We conducted experiments on the JFLEG test set to examine the effect of the number of training examples on the model's performance. 
The task instruction was kept consistent across all experiments. 
To perform the experiments, we randomly sampled examples from the training set of the JFLEG dataset. 
We tested the model with 2, 4, 8, 16, 32, and 64 examples, limiting the maximum number of examples to 64 due to the maximum input length of the model employed in our study.

\subsubsection{Result}

The results obtained from each experimental setting are presented in Figure~\ref{fig:num-example}. 
Our experiments revealed a clear trend: performance improved as the number of examples increased. 
Our analysis further indicated that the models benefit from having more examples during the few-shot learning process. 
The highest score of 69.25 was achieved with 64 examples, suggesting that providing more examples can offer better guidance and context for the models to understand and effectively perform the task.

However, it is important to note that performance improvement is not strictly linear with the increase in the number of examples. 
For instance, the score slightly dipped from 67.11 to 66.67 when the number of examples increased from 16 to 32. 
This deviation from linearity could be attributed to the quality of the examples or the inherent variability in the models' performance. 
Further investigation is required to understand better the factors contributing to these fluctuations and identify the optimal number of examples needed to maximize performance.

\section{Controllability through Prompt}
\label{sec:controllability}

In this section, we explore GPT-3's controllability for GEC tasks through prompt-based methods. 
Our experiments focus on two settings: 
(1) comparing the model's performance when instructed to make minimal edits versus emphasizing fluency, and 
(2) tailoring the editing to different learner levels, including beginner, intermediate, advanced, and native speakers. 
We aim to gain insights into GPT-3's flexibility and controllability under various conditions.
We also analyze the relative influence of task instruction and examples to identify the factor that significantly impacts the model's output.

\subsection{Minimal vs. Fluency Edits}
\label{exp-control-fluency}
\subsubsection{Settings}

We evaluated controllability for minimal and fluency edits using the CoNLL2014 and JFLEG test sets, respectively. 
CoNLL2014 is a widely-used benchmark for GEC tasks, while JFLEG focuses on fluency-based evaluation.
We conducted experiments in zero-shot and 16-shot settings.
We used different task instructions to control the settings in the prompts, such as '\texttt{Revise the following sentence with proper grammar}' for minimal edits and '\texttt{Revise the following sentence to improve fluency}' for fluency edits.

We assessed the models using performance-based evaluation and edit distance-based evaluation. 
Performance-based evaluation measures the model's error correction or fluency improvement ability, while edit distance-based evaluation quantifies the difference between original and revised sentences, offering insights into the extent of editing performed.

\subsubsection{Results}

\paragraph{Performance-based Evaluation}

Figure~\ref{fig:control-fluency-performance-all} compares scores in performance-based evaluation for minimal and fluency edit instructions. 
In the zero-shot setting, minimal edit instructions perform better on the CoNLL2014 test set, while both instructions yield comparable scores on the JFLEG set. 
In the few-shot setting, higher scores are observed when using corresponding task instructions for each test set, emphasizing the effectiveness of text prompts in controlling editing settings. 
The discrepancy between zero-shot and few-shot settings might be due to the model's limited understanding of the task in the zero-shot setting. 
Additional examples in the few-shot setting enable the model to comprehend the task's objective better and adjust its output accordingly.

Additionally, we also compared the prompt-based method with a supervised controlling method that uses special tokens as in \newcite{johnson-etal-2017-googles}, where different special tokens were used to control target languages in multilingual translation. 
We trained a Transformer (Big) encoder-decoder with annotated data tagged with special tokens indicating minimal and fluency edits settings. 
Despite using more training data, this supervised method failed to control specific settings while achieving higher scores on both test sets with fluency edit tokens, as in Figure~\ref{fig:control-fluency-performance-all}. 
This finding highlights the potential advantages of the prompt-based approach.

\begin{figure}[t]
    \centering
    \includegraphics[width=1.0\linewidth]{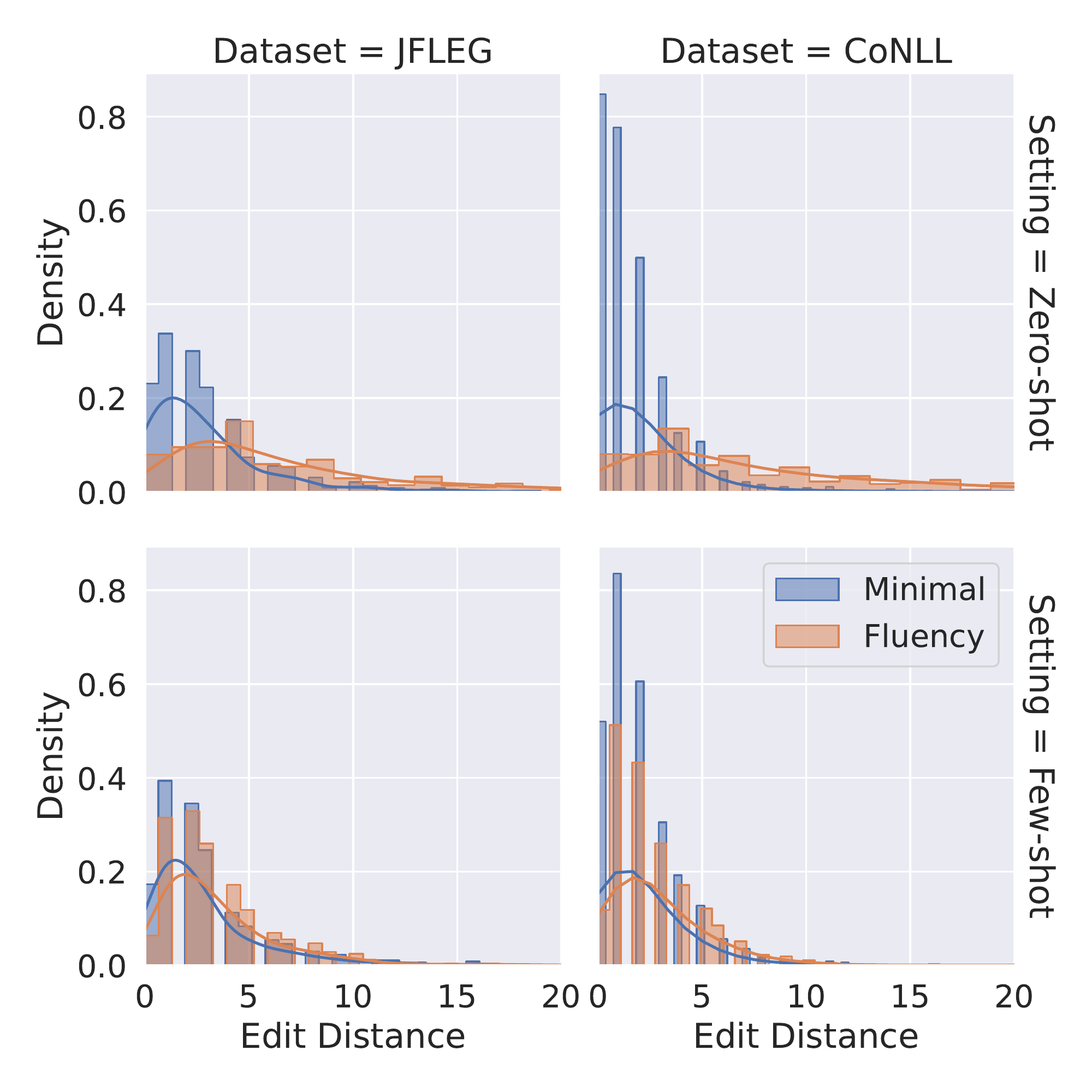}
    \caption{Edit distance distributions for minimal and fluency edits on CoNLL2014 and JFLEG test sets, respectively, as part of the edit distance-based evaluation for controllability of prompts.}
    \label{fig:control-fluency-editdistance-all}
\end{figure}

\paragraph{Edit Distance-based Evaluation}

Figure~\ref{fig:control-fluency-editdistance-all} presents edit distance distributions for each setting as part of edit distance-based evaluation. 
A shift to the right indicates more edits performed with fluency edit instructions. In the few-shot setting, the difference in edit distance distributions between minimal and fluency edits is smaller than in the zero-shot setting, which can be attributed to the influence of the examples presented in the prompt.
The model's ability to generalize from examples in the few-shot setting may diminish the difference in edit distance between the two settings, further emphasizing the importance of carefully selected examples.

In summary, the prompt-based method using GPT-3 can effectively control GEC task outputs for either minimal or fluency edits. 
Controllability is more evident in few-shot settings, where additional examples help the model adapt its behavior according to the given instructions. 
The edit distance-based evaluation further supports the model's ability to adjust its editing behavior based on the prompt, showcasing its potential for practical applications.

\begin{figure}[t]
    \centering
    \includegraphics[width=\linewidth]{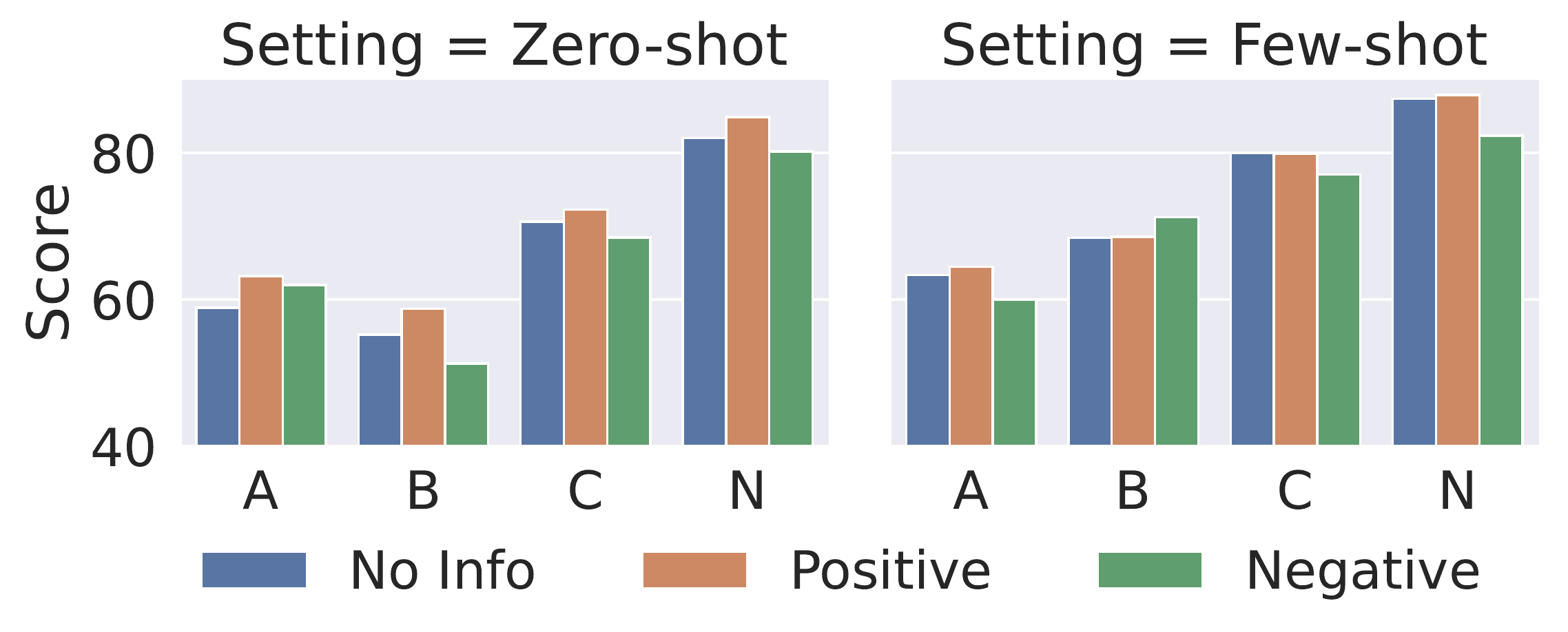}
    \caption{Impact of task instructions with varying additional information on GPT-3's performance in GEC tasks, evaluated on the validation sets of W{\&}I+LOCNESS. The experiment features three settings: No Info, Positive Info, and Negative Info. The x-axis represents different CEFR levels (A, B, C) and native speakers (N) included in the validation set.}
    \label{fig:control-learner-performance-all}
\end{figure}

\subsection{Learner Level-based Correction}
\label{exp-learner-level}
\subsubsection{Settings}

In this section, we examine GPT-3's adaptability to diverse GEC task requirements and contexts by analyzing the impact of varying additional information in task instructions.
We conducted experiments in both zero-shot and few-shot (16-shot) settings.
We utilized the W{\&}I+LOCNESS validation sets, comprising text from various CEFR levels (A: Beginner, B: Intermediate, C: Advanced) and native speakers (N) as evaluation sets.
We devised an experiment with three settings based on the following types of additional information (refer to Appendix~\ref{sec:appendix-learner}):

\paragraph{No Info:} No extra information is provided.
\paragraph{Positive Info:} Information that supports the input sentence's characteristics, such as the number of errors to be revised. 
Example: "\texttt{Revise mistakes in the following text written by a beginner learner with a lot of mistakes.}"
\paragraph{Negative Info:} Information that contrasts with the input sentence's characteristics, e.g., a text written by a beginner learner with many errors but described as having few. 
Example: "\texttt{Revise mistakes in the following text written by an advanced learner with only a few mistakes.}"

\subsubsection{Results}

Figure~\ref{fig:control-learner-performance-all} shows the results of controlling task instruction with additional information on learner levels.
In the zero-shot setting, positive information improved performance, while negative information adversely impacted output across most learner levels.
This demonstrates the influence of additional information in task instructions.
In the few-shot setting, task instructions without additional information (No Info) achieved higher scores than cases with Positive Info, suggesting that the model effectively utilizes examples to understand the desired correction level.
However, with Negative Info, performance dropped for most learner levels compared to No Info and Positive Info cases.

\subsection{Effect of Task Instruction vs. Examples}
\label{exp-task-example}

In this section, we present an experiment to examine the relative effect of task instruction and examples on GPT-3's performance in controllability, in few-shot settings. 
Our primary objective is to determine which of these two components, task instruction and example, has a more significant impact on the model's outputs. 
Moreover, we extend our investigation to explore the influence of examples on the editing process of the output, providing a more comprehensive understanding of the interplay between these variables in the context of few-shot learning.

\subsubsection{Settings}
To investigate the relative influence of task instructions and examples independently, we designed two experiments, each featuring distinct conditions:

\paragraph{Varied Task Instruction with Fixed Examples (VIFE)} 
We modified the task instructions while maintaining a constant set of examples. 
This approach allows us to assess the influence of task instructions on the model's performance.

\paragraph{Fixed Task Instruction with Varied Examples (FIVE)} 
We utilized a single task instruction and altered the set of examples. 
This condition helps us evaluate the impact of examples on the model's performance.

In this experiment, we employed the JFLEG and CoNLL2014 test sets. 
We assessed the performance using ${\rm F}_{0.5}$ score for CoNLL2014 and GLEU for JFLEG. 
For the VIFE condition, we prepared a fixed set of examples and a varied set of task instructions for each dataset, similar to the approach in Section \ref{exp-control-fluency}.
We used task instructions that requested the model to perform minimal edits on the CoNLL2014 test set and fluency edits on the JFLEG test set. 
For the FIVE condition, we prepared fixed task instructions and varied examples from the training sets of NUCLE and JFLEG, which correspond to minimal and fluency edits, respectively.
We conducted experiments in this section with 16-shot setting.

\begin{figure}[t]
    \centering
    \includegraphics[width=\linewidth]{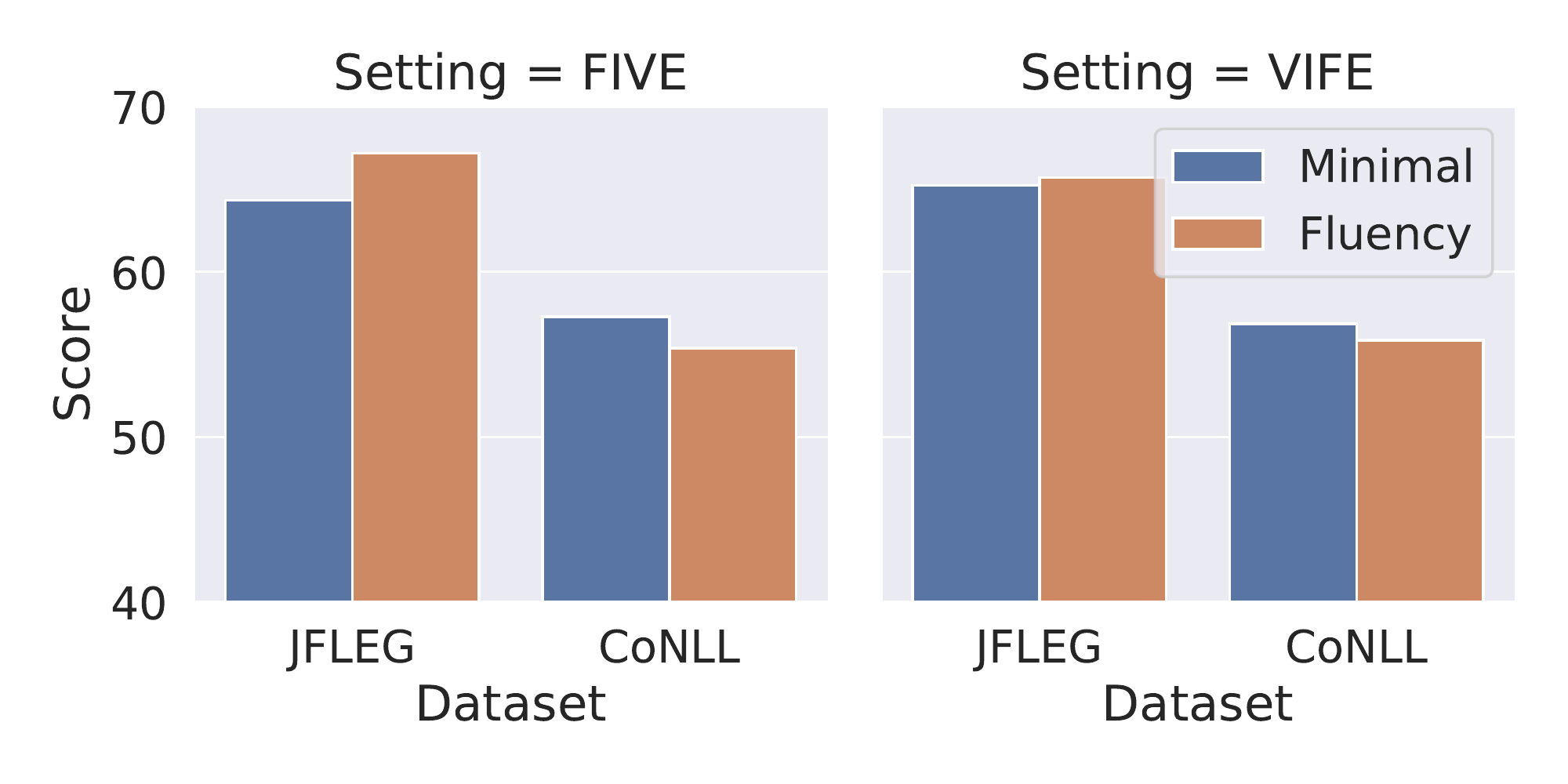}
    \caption{Comparison of the impact of task instructions and number of examples in few-shot settings. VIFE condition examines the effect of varied task instructions with fixed examples, while FIVE condition evaluates the impact of fixed task instructions with varied examples.}
    \label{fig:compare-instruction-example}
\end{figure}

\subsubsection{Results}

Figure~\ref{fig:compare-instruction-example} summarizes the results regarding the performance scores.
In both CoNLL2014 and JFLEG, we observed performance gaps between the two settings, minimal and fluency edits.
However, the gaps were more drastic when changing the example set compared to varying the task instruction.
These results suggest that examples play a more critical role in controlling the model's behavior than task instructions, as changing the example set leads to more significant differences in achieving the desired output.
This is likely because examples provide specific and contextual information, while task instructions can be abstract and open to interpretation.
This highlights the importance of carefully selecting examples to optimize model performance.

\begin{table}[t]
    \centering
    \begin{tabular}{lcc}
        \hline
        \multirow{2}{*}{Test set}& \multicolumn{2}{c}{Example from}\\
        & JFLEG & NUCLE \\
        \hline
        & \multicolumn{2}{c}{Fluency Edits} \\
        JFLEG & \textbf{0.1569} & 0.1893 \\
        CoNLL2014 & 0.4443 & \textbf{0.4058} \\
        \hline
        & \multicolumn{2}{c}{Minimal Edits} \\
        JFLEG & \textbf{0.2283} & 0.3038 \\
        CoNLL2014 & 0.4158 & \textbf{0.3768} \\
        \hline
    \end{tabular}
    \caption{Impact of example set on GPT-3's performance in few-shot settings evaluated on JFLEG and CoNLL2014 test sets, measured by Jensen-Shannon distance. Diagonal entries show closer alignment between model output and corresponding example set.}
    \label{tab:control-examples}
\end{table}

We further investigated the example set's impact on model output, using Jensen-Shannon distance to compare edit distance distributions in both minimal and fluency edits settings. 
Lower Jensen-Shannon distance indicates a more similar edit distribution between the example set and model output.
Results in Table~\ref{tab:control-examples} show lower distances in diagonal entries, signifying closer alignment between the model output and corresponding example set. 
This highlights the importance of carefully selecting examples to guide the model in generating outputs with desired characteristics.

\section{Related Work}

Supervised learning approaches have predominantly driven GEC research, resulting in state-of-the-art performance. 
Encoder-decoder models are commonly employed in GEC using supervised learning. 
\citet{yuan-briscoe-2016-grammatical} first applied an encoder-decoder model to GEC, inspiring subsequent researchers to propose various encoder-decoder-based GEC models~\cite{ji-etal-2017-nested,chollampatt-ng-2018-multilayer,junczys-dowmunt-etal-2018-approaching,zhao-etal-2019-improving,kaneko-etal-2020-encoder,yamashita-etal-2020-cross}. 
These methods typically rely on large training datasets containing parallel sentences with and without grammatical errors~\cite{kiyono-etal-2019-empirical}. 
However, scalability remains challenging, as labeled data is required for each specific situation, such as grammar correction style or input text domain.

Unsupervised GEC approaches aim to reduce dependency on labeled data by leveraging unsupervised learning techniques, including PLMs, hand-crafted rules, denoising autoencoders, or unsupervised machine translation~\cite{grundkiewicz-etal-2019-neural,grundkiewicz-junczys-dowmunt-2019-minimally,flachs-etal-2019-noisy,solyman2021synthetic,koyama-etal-2021-various}. 
However, these methods necessitate creating large-scale pseudo data for model training, making it difficult to generate pseudo-data and train models for different learning scenarios. 
Some studies have proposed unsupervised GEC methods using PLMs~\cite{alikaniotis-raheja-2019-unreasonable,yasunaga-etal-2021-lm}, but they have not focused on prompt-based methods with PLMs.

Recently, the GPT-3 model~\cite{brown-gpt3-2020} has demonstrated remarkable performance across various NLP tasks, although its GEC performance remains limited. 
\newcite{schick2022peer} employed a simple zero-shot prompt for GEC, while \newcite{DwivediYu2022EditEvalAI} conducted a more comprehensive analysis using diverse zero-shot prompts. 
\newcite{coyne2023analysis} and \newcite{fang2023chatgpt} compared the latest GPT-3 model's performance (text-davinci-003) and ChatGPT against GEC leaderboard models and reference edits, finding that these prompt-based methods exhibited strong GEC performance. 
However, automatic metrics and human evaluations occasionally disagreed on the relative quality of corrections.

Controlling GEC model generation is crucial but remains underexplored. 
\citet{hotate-etal-2019-controlling} proposed a GEC method that controls the degree of correction by tagging input with the correction level, but it requires supervised learning with parallel data. 
Additionally, \citet{hotate-etal-2020-generating} suggested a beam search method to control GEC correction diversity by dynamically updating search tokens within the beam based on the likelihood of predicting source sentence tokens. 
While this method enables model control without additional training, it falls short in accommodating specific learner requests, such as minimal and fluency edits.

GEC model evaluation methods have been proposed based on learner levels and correction styles. 
To account for differences in correction styles and domains, \citet{maeda-etal-2022-impara} introduced a method to train evaluation models using only parallel data. 
\citet{takahashi-etal-2022-proqe} created proficiency-annotated data to train evaluation models and developed an evaluation method that considers proficiency by fine-tuning PLMs~\cite{yoshimura-etal-2020-reference}.

\section{Conclusion}
In conclusion, this study demonstrates the potential of using prompt-based methods with GPT-3 for GEC tasks, achieving competitive performance compared to traditional supervised and unsupervised methods. 
By carefully crafting task instructions and examples, we show that GPT-3 can be effectively controlled to focus on different aspects of the GEC process and adapt to diverse learning needs. 
Our findings highlight the importance of optimizing task instructions and example selection to enhance the performance and controllability of GPT-3, paving the way for further research on refining prompt engineering techniques and exploring their applicability to other NLP tasks and language models.

\section{Educational Implications and Community Benefits}
Our study provides valuable implications for education. 
The controllability of large-scale language models in GEC tasks can be leveraged to design personalized language instruction. 
It allows educators to provide feedback that matches individual students' proficiency levels and focuses on specific areas for improvement. 
For learners, instant, tailored feedback can enhance their language learning process. 
Moreover, our findings can improve intelligent tutoring systems, making them more responsive to individual needs. 
Beyond education, our research can enhance language-based interfaces and AI communication systems, offering more accurate and context-specific language corrections. 
This study lays the groundwork for future exploration into how large language models can improve language education and literacy.

\section{Limitation}
While our study provides valuable insights into the use of prompt-based methods with GPT-3 for GEC tasks and its controllability, several limitations should be acknowledged.

\paragraph{Focus on GPT-3:} This study exclusively examines GPT-3 as the language model for GEC tasks. 
While GPT-3 has shown remarkable performance in various NLP tasks, other pre-trained language models, such as GPT-4, may offer different results. 
A broader investigation that includes other language models would provide a more comprehensive understanding of the applicability of prompt-based methods in GEC tasks.

\paragraph{Limited evaluation metrics:} The evaluation of GPT-3's performance and controllability in our experiments mainly relies on quantitative metrics, such as edit distance and task scores.
These metrics may not fully capture the nuances of grammatical error correction or the model's ability to adapt to different learning scenarios. 
Additional qualitative analysis, along with more diverse evaluation metrics, could provide a richer understanding of the model's performance and controllability.

\paragraph{Variability in examples:} While our study highlights the importance of example selection in few-shot settings, we do not thoroughly explore the impact of example quality or diversity. 
The effect of using different types of examples or a more diverse set of examples remains to be investigated, which could further inform the design of effective example sets for prompt-based GEC tasks.
By addressing these limitations in future research, we can further advance our understanding of the performance and controllability of prompt-based methods with GPT-3 and other language models in GEC tasks and beyond.

\paragraph{Potential fine-tuning on test data:} There is a possibility that GPT-3 has been fine-tuned (instruction tuning) on the test data we are using, which might explain the higher evaluation scores compared to previous research. 
As this information has not been disclosed, we are unable to verify it at this time. 
This point should be taken into consideration when interpreting our results.

\section*{Acknowledgements}
These research results were obtained partially from the commissioned research (No. 225) by National Institute of Information and Communications Technology (NICT), Japan.

\bibliography{custom}
\bibliographystyle{acl_natbib}

\newpage
\appendix
\section{Prompts for Investigation on Instruction Effect}
\label{sec:appendix-instruction}

All instructions used for experiments described in Section~\ref{exp-instruction-selection} are listed in Table~\ref{tab:all-instructions}.

\begin{table*}[h]
  \centering
  \begin{tabularx}{\textwidth}{XX}
    \hline
    \textbf{Type} & \textbf{Task Instruction} \\
    \hline
    \midrule
    Instructive & Correct grammatical errors in this sentence \\
    & Revise grammatical mistakes in the following text. \\
    & Edit this paragraph for grammar mistakes. \\
    & Find and fix any errors in this sentence. \\
    & Rewrite this sentence to correct its grammar. \\
    & Identify and correct the grammar errors in this text. \\
    & Make any necessary grammar corrections to this passage. \\
    & Correct the grammar in this sentence without changing its meaning. \\
    & Find and correct the errors in this paragraph. \\
    & Proofread this text and correct any grammar mistakes. \\
    \hline
    Misleading & Paraphrase the following sentence. \\
    & Rewrite the following text to make it clearer.\\
    & Revise this paragraph to improve its clarity.\\
    & Clarify the meaning of this sentence by rephrasing it.\\
    & Make this sentence more concise without changing its meaning.\\
    & Improve the readability of this text by rewording it.\\
    & Reconstruct this sentence to enhance its clarity.\\
    & Paraphrase this text to make it more comprehensible.\\
    & Rewrite this paragraph to convey the same information in a clearer way.\\
    & Edit this sentence to improve its coherence and flow.\\
    \hline
    Irrelevant & Translate the following sentence in to Japanese.\\
    & Write a news headline about this sentence.\\
    & Create a meme based on the following text.\\
    & Write a short story based on this sentence.\\
    & Compose a poem using the words in this paragraph.\\
    & Write a summary of this text.\\
    & Analyze the use of metaphor in this sentence.\\
    & Explain the historical context of this passage.\\
    & Write a tweet about this text.\\
    & Write a letter to your future self based on the following sentence.\\
    \hline
  \end{tabularx}
  \caption{Prompts for Instruction Effect Investigation, showing three types of task instructions with ten candidate prompts each. The types include Instructive, Misleading, and Irrelevant prompts.}
  \label{tab:all-instructions}
\end{table*}

\section{Prompts for Learner's Level-based Control}
\label{sec:appendix-learner}

All instructions and additional information used for experiments described in Section~\ref{exp-learner-level} are listed in Table~\ref{tab:learner-info}.

\begin{table*}[h]
  \centering
    \begin{tabularx}{\textwidth}{XX}
    \hline
    Info     & Task Instruction \\
    \hline
    \textbf{Beginner} & \\
    \hline
    No Info          & Revise mistakes in the following text \\
    Positive Info     & Revise mistakes in the following text written by a beginner learner with a lot of mistakes \\
    Negative Info         & Revise mistakes in the following text written by an advanced learner with only a few mistakes \\
    \hline
    \textbf{Intermediate} & \\
    \hline
    No Info        & Revise mistakes in the following text \\
    Positive Info     & Revise mistakes in the following text written by an intermediate learner with some mistakes \\
    Negative Info     & Revise mistakes in the following text written by a native speaker \\
    \hline
    \textbf{Advanced} & \\
    \hline
    No Info       & Revise mistakes in the following text \\
    Positive Info     & Revise mistakes in the following text written by an advanced learner with only a few mistakes \\
    Negative Info     & Revise mistakes in the following text written by a beginner learner with a lot of mistakes \\
    \hline
    \textbf{Native} & \\
    \hline
    No Info          & Revise mistakes in the following text \\
    Positive Info      & Revise mistakes in the following text written by a native speaker \\
    Negative Info      & Revise mistakes in the following text written by a beginner learner with a lot of mistakes \\
    \hline
    \end{tabularx}
  \caption{All prompts used in experiments investigating the controllability of learner level-based edits.}
  \label{tab:learner-info}
\end{table*}

\end{document}